\documentclass[sigconf]{acmart}

\usepackage{algorithm}
\usepackage{algpseudocode}  
\usepackage{multicol} 
\usepackage{graphicx}
\usepackage{multirow}
\usepackage{subcaption}
\usepackage{xcolor}
\usepackage{natbib}  
\usepackage{newfloat}   
\usepackage{listings}

\usepackage{enumitem}

\usepackage[normalem]{ulem}     
\usepackage[switch]{lineno}     

\useunder{\uline}{\ul}{}
\newcommand{\parm}{\mathord{\color{black!33}\bullet}}

\AtBeginDocument{%
  \providecommand\BibTeX{{%
    \normalfont B\kern-0.5em{\scshape i\kern-0.25em b}\kern-0.8em\TeX}}}

\copyrightyear{2024} \acmYear{2024} \setcopyright{acmlicensed}\acmConference[KDD '24]{Proceedings of the 30th ACM SIGKDD Conference on Knowledge Discovery and Data Mining}{August 25--29, 2024}{Barcelona, Spain} \acmBooktitle{Proceedings of the 30th ACM SIGKDD Conference on Knowledge Discovery and Data Mining (KDD '24), August 25--29, 2024, Barcelona, Spain} \acmDOI{10.1145/3637528.3672069} \acmISBN{979-8-4007-0490-1/24/08}

\begin{document}


\title{Masked LoGoNet: Fast and Accurate 3D Image Analysis for Medical Domain}

\author{%
  \textbf{Amin Karimi Monsefi\textsuperscript{§}, 
  Payam Karisani\textsuperscript{†}, 
  Mengxi Zhou\textsuperscript{§}, Stacey Choi\textsuperscript{§}}, \\
  \textbf{Nathan Doble\textsuperscript{§},
  Heng Ji\textsuperscript{†},
  Srinivasan Parthasarathy\textsuperscript{§},
  Rajiv Ramnath\textsuperscript{§}} \\
  \texttt{\{karimimonsefi.1, ramnath.6\}@osu.edu}, 
  \texttt{karisani@illinos.edu} \\
  \textsuperscript{§}The Ohio State University, 
  \textsuperscript{†}University of Illinois at Urbana-Champaign, 
}

\renewcommand{\shortauthors}{Amin Karimi Monsefi et al.}

\begin{abstract}

Standard modern machine-learning-based imaging methods have faced challenges in medical applications due to the high cost of dataset construction and, thereby, the limited labeled training data available. Additionally, upon deployment, these methods are usually used to process a large volume of data on a daily basis, imposing a high maintenance cost on medical facilities. In this paper, we introduce a new neural network architecture, termed LoGoNet, with a tailored self-supervised learning (SSL) method to mitigate such challenges. LoGoNet integrates a novel feature extractor within a U-shaped architecture, leveraging Large Kernel Attention (LKA) and a dual encoding strategy to capture both long-range and short-range feature dependencies adeptly. This is in contrast to existing methods that rely on increasing network capacity to enhance feature extraction. This combination of novel techniques in our model is especially beneficial in medical image segmentation, given the difficulty of learning intricate and often irregular body organ shapes, such as the spleen. Complementary, we propose a novel SSL method tailored for 3D images to compensate for the lack of large labeled datasets. Our method combines masking and contrastive learning techniques within a multi-task learning framework and is compatible with both Vision Transformer (ViT) and CNN-based models. We demonstrate the efficacy of our methods in numerous tasks across two standard datasets (i.e., BTCV and MSD). Benchmark comparisons with eight state-of-the-art models highlight LoGoNet's superior performance in both inference time and accuracy. Code available at: \href{https://github.com/aminK8/Masked-LoGoNet}{https://github.com/aminK8/Masked-LoGoNet}.

\end{abstract}




\maketitle

\small
\section{Introduction}

Accurate medical image segmentation can facilitate disease diagnosis and treatment planning \cite{hatamizadeh2021swin, zhou2022using}. One of the fundamental difficulties in this task is the presence of organs or structures that span a large receptive field. These structures may have irregular shapes, complex boundaries, or significant variations in appearance, making the segmentation task particularly demanding. Additionally, the high cost of expert annotation in this domain restricts the availability of large-scale labeled datasets. Consequently, it limits the applicability of general domain computer vision methods \cite{haghighi2022dira, cai2023dual}. Furthermore, deployed systems usually process a large volume of images on a daily basis, which demands a substantial computational resources and leaves a large carbon footprint \cite{li2021multi}. In the present work, we propose a fast and accurate image segmentation architecture for the medical domain. We also propose a pre-training algorithm to exploit unlabeled images, and therefore, alleviate the demand for human annotation. 

Our architecture is based on the widely adopted U-shaped model. We particularly employ two strategies to enhance the inference speed, and simultaneously, maintain the prediction accuracy. First, in contrast to existing models that rely on Convolutional Neural Networks (CNNs) and Vision Transformers (ViTs) as encoders \cite{dosovitskiy2020image, isensee2018nnu}, we employ the large-kernel attention model (LKA) \cite{guo2022visual} in our feature extractor, which we term \textbf{ULKANet} (\textbf{U}net \textbf{L}arge \textbf{K}ernel \textbf{A}ttention \textbf{Net}work). As we discuss in the next section, CNN and ViTs-based models suffer from a high memory complexity, are slower during inference, and lack a proper strategy to process image sequences.\footnote{The term "sequence" in 3D medical imaging refers to a series of volumetric data that can be either a temporal sequence, capturing changes over time in a specific anatomical region, or a spatial sequence, consisting of different slices from a 3D volume to provide a comprehensive view of the anatomy from various angles.} On the other hand, our method is demonstrably more efficient due to the presence of LKA in the encoder.

Our second strategy is to enhance feature extraction through an inductive bias. Learning short-range and long-range dependencies is essential in medical image segmentation due to the large receptive field of organs. Existing studies employ U-Net with the attention mechanism, and vertically scale up their architecture to increase the network capacity for handling feature dependencies \cite{oktay2018attention, cai2022ma}. In contrast to these methods, we incorporate our encoder (ULKANet) into a dual encoding algorithm to learn local (short-range) as well as global (long-range) features. This enables us to keep the network size manageable, and at the same time, maintain the prediction accuracy. We term this model \textbf{LoGoNet} (\textbf{Lo}cal and \textbf{G}l\textbf{o}bal \textbf{Net}work)\footnote{This work was supported by an NSF MRI Grant $\#2018627$}. 
Our model is particularly advantageous for segmenting organs such as the spleen, which has an elongated shape and irregular corners. Such body organs demand the extraction of global and local features for segmentation.



Finally, we propose a novel self-supervision technique for 3D images to address the lack of labeled training data. Our self-supervision method combines masking and multi-task learning. Using a multi-clustering algorithm, we generate a list of pseudo-labels for each unlabeled image. We then methodically mask selected parts of these images to implicitly feed the structural information of the unlabeled data into our model. A property of our proposed SSL technique lies in its versatility, as it seamlessly supports both CNN and ViT-based models. This flexibility sets our strategy apart from conventional SSL approaches, which often cater to a specific architecture \cite{he2022masked, zhou2021ibot, kakogeorgiou2022hide}. Furthermore, our strategy leverages the inherent characteristics of 3D medical images, specifically embracing the concept of sequential images and neighborhood information of voxels in 3D images. 

We evaluate our techniques on numerous tasks across two datasets, i.e., the BTCV dataset \cite{gibson2018automatic} for segmenting body organs, and the MSD dataset \cite{simpson2019large} that encompasses diverse tasks in medical imaging, ranging from liver tumors to cardiac and lung segmentation. Additionally, we benchmark our method against eight state-of-the-art baseline models. The results demonstrate the effectiveness and efficiency of our techniques. To offer a thorough insight into the attributes of our approach, we undertook extensive experiments, meticulously showcasing our model's features and capabilities.
To summarize, our contributions are threefold:
\begin{itemize}[noitemsep,topsep=0pt,leftmargin=*]
    \item We propose a resource-efficient model based on the commonly used U-shaped architecture. Our model has a short inference time and, at the same time, outperforms state-of-the-art methods. We achieve this by employing two strategies: first, instead of relying on CNN or ViT-based techniques, we utilize the large-kernel attention method to reduce computational complexity. Second, instead of vertically scaling up our network to improve feature extraction, we use a dual encoding algorithm to facilitate the task. We empirically demonstrate that our strategies combined achieve the best inference time and the highest precision.
    \item We propose a multi-task self-supervision technique to exploit unlabeled images, and to overcome the lack of labeled data by employing a new masking approach specifically designed for 3D images.
    \item We evaluate the efficacy of our model on numerous tasks across two datasets, and show that it outperforms eight state-of-the-art baseline models.
\end{itemize}

\section{Related Work}

To model long-range dependencies in images, existing studies mostly use vision transformers \cite{cao2023swin, hatamizadeh2021swin, valanarasu2021medical, karimi2023crashformer, wu2022fat, he2021global, perera2024segformer3d, azad2023foundational}, and draw ideas from sequence modeling in Natural Language Processing~(NLP). A limitation of these approaches is their treatment of images as 1D sequences, thereby overlooking the input's inherent 2D or 3D structure. They struggle to grasp the spatial relationships between pixels, leading to poor performance in tumor detection or organ segmentation tasks.
Additionally, they suffer from quadratic memory complexity, leading to high processing costs and slowness for high-resolution images, especially in the 3D context \cite{singh20203d, khan2022transformers, jangid2024q, wang2023adaptive}. In contrast, our proposed model, ULKANet, adopts an attention mechanism with LKA\footnote{LKA \cite{guo2022visual} is a method for computer vision tasks that effectively captures long-range relationships from input features. LKA reduces computational costs while generating attention maps highlighting essential features without additional normalization functions by decomposing large kernel convolutions into spatial local, long-range, and channel convolutions. } to handle long-range dependencies while preserving the spatial structure of the images. This distinctive property enables our model to capture spatial patterns of the input more effectively, resulting in more informative representations. This is particularly advantageous in detecting tumors, where the conditions may extend over a considerable area, and models that rely solely on local features often fail to detect such cases \cite{wang2022medical}.

In addressing dependencies within data, various techniques are employed based on the range of the dependencies. CNN-based models have proven effective for short-range dependencies, leveraging convolutional operations to identify relevant spatial patterns efficiently. Through this approach, hierarchical representations are learned, enhancing the understanding of the intrinsic structure of the data \cite{xia20203d, isensee2018nnu, zhou2022structure}. However, our methodology takes a comprehensive approach, recognizing the importance of long and short-range dependencies. We adopt a dual encoding strategy to achieve this, incorporating an attention mechanism in parallel mode. This dual encoding technique enables the simultaneous capture and encoding of both types of dependencies, providing a more holistic representation of the underlying relationships in the data.

Next, the lack of labeled training data is a primary challenge in medical image analysis. To address this challenge, some studies have focused on domain-specific pretext tasks, as seen in \citet{zhu2020rubik, zhao2021anomaly, cao2020auto, he2021autoencoder}, and, \citet{xu2021deformed2self}. Others, such as \citet{zhou2020comparing}, adapt contrastive learning techniques to suit medical data by focusing on feature level contrasts, creating homogeneous and heterogeneous data pairs by mixing image and feature batches, and utilizing a momentum-based teacher-student architecture. A comprehensive evaluation of various SSL strategies for 3D medical imaging was conducted by \citet{taleb20203d}. \citet{azizi2021big} demonstrated the benefits of pre-training a model on ImageNet for dermatology image classification, showcasing the potential of transfer learning in the medical imaging domain.

\section{Proposed Model}

Figure \ref{fig:overview} illustrates the architecture of our model LoGoNet. The forward pass begins by processing the input data in parallel. We have two modules in this stage, the global and the local modules. In the global module, the original data cube\footnote{"Cube" typically refers to a three-dimensional (3D) region of interest (ROI) within the volumetric medical image. Medical images, such as those obtained from MRI or CT scans, are often represented as 3D volumes, where each voxel (3D pixel) contains intensity or other information about the tissue or structures being imaged. A cube in this scenario is a 3D subset of the entire image volume.} is fed into our feature extractor (ULKANet). In the local module, the same data cube is partitioned into smaller cubes, and then, each cube is processed by a separate feature extractor. Afterwards, the resulting feature tensors are concatenated to reconstruct the input. Then the outputs of the global and the local modules are aggregated by an element-wise summation operator--note that they have the same dimensions. Finally, the resulting tensor is passed through a convolution kernel followed by a 3D batch normalization operator and a GELU activation function to shape the input to our final classifier. Our final classifier is a convolution kernel.

In the next section, we discuss our $3D$ encoder-decoder architecture (ULKANet), which is armed with a $3D$ adaptation of LKA in the encoding phase. We then explain our local-global dual encoding strategy, which enables our model to extract feature dependencies at varying scales. After describing our model in detail in Sections \ref{sub-sec:ulkanet} and \ref{sub-sec:dual-enc}, we then explain our novel pre-training method in Section \ref{sub-sec:pretrain}. We use the pre-training algorithm to initialize the parameters of our model before beginning to fine-tune the network on labeled data.

\subsection{LKA in Feature Extractor: An Alternative to CNN and ViTs-based Models} 
\label{sub-sec:ulkanet}

\begin{figure}[ht!]
    \centering
    \begin{subfigure}[t]{0.50\linewidth}
        \centering
        \includegraphics[width=\linewidth]{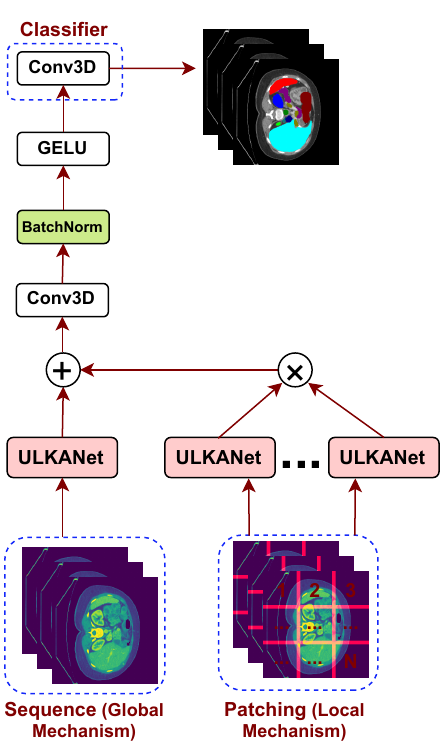}
        \caption{Proposed Model}
        \label{fig:overview}
    \end{subfigure}~~
    \begin{subfigure}[t]{0.48\linewidth}
        \centering
        \includegraphics[width=\linewidth]{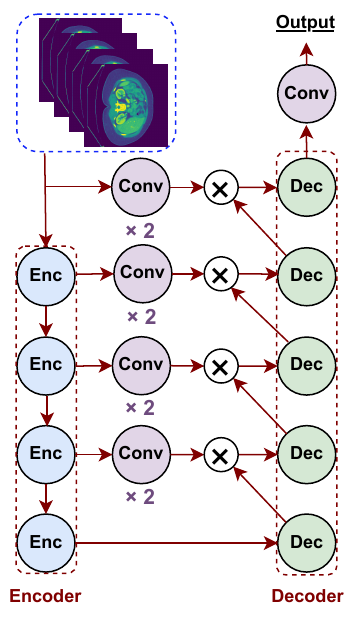}
        \caption{ULKANet}
        \label{fig:ulkanet}
    \end{subfigure}
    \caption{\textbf{\ref{fig:overview})} Overview of our model LoGoNet. In order to take into account the local and global feature dependencies in images, they are fed into the model in parallel. In the local mechanism, the input data is partitioned into small parts, and each part is separately fed into our feature extractor (ULKANet). \textbf{\ref{fig:ulkanet})} Overview of the ULKANet Architecture. A U-shaped network with the encoder-decoder design. Blue circles represent encoder blocks, and green circles represent the decoder blocks. The $+$ sign represents element-wise summation, and the $\times$ sign represents the concatenation operator.}
\end{figure}

Figure \ref{fig:ulkanet} illustrates an overview of our feature extractor (ULKANet), which is a U-shaped model and has an encoder and a decoder. The encoder consists of a sequence of blocks. Each block consists of a repeating sequence of three components: a patch embedding component, a chain of transformer-like modules that employ LKA ($L_i$ modules for $i^{th}$ block of the encoder), and a layer normalization component. For conciseness, Figure \ref{fig:ulkanet} only shows the top-level blocks, while a detailed illustration of the model architecture and inner components is provided in the appendix section \ref{appendix_ulkanet}.

The Patch Embedding component plays a crucial role in the processing of input data within the encoder block, transforming the input into a tensor that is subsequently passed to the next component in the sequence. Throughout the current encoder block, the dimension of the embedding vectors remains constant, denoted as $dim$. The mathematical representation of the projection operation is defined as follows:
\begin{equation}
Patch = Norm(Conv3D(X, dim, k, padding=\frac{k}{2})).flatten(2), 
\label{eq:patch}
\end{equation}

\noindent where $X$ represents the input with five dimensions $(b, C, \text{seq}, H, W)$, and $b$ is the batch size, $C$ is the channel size, $k$ is the size of the 3D convolution kernel, $dim$ is the number of channels for the output of $\text{Conv3D}$, and $\text{Norm}$ represents the batch normalization operator. $(\text{seq}, H,  W)$ denotes the size of the 3D input, and the $\text{flatten}$ operation results in a tensor with dimensions $(b, dim, \text{seq} \times H \times W)$. The Patch Embedding process serves to efficiently capture and represent the relevant features of the input data, facilitating the subsequent stages of the network architecture.

To enable our model to efficiently extract complex feature dependencies that are often present in medical images, we opt for using transformer modules. However, instead of using the regular transformers with self-attention that is slow and needs more memory \cite{singh20203d}, we use LKA \cite{guo2022visual} in the attention layer. This type of attention mechanism decomposes large convolution kernels into spatial dependencies and channel convolutions. It enables our model to go deeper and remain memory efficient. The attention module is implemented as follows:
\begin{equation}
\textit{Atts} = \textit{Conv3D}_{1\times 1} \left( \textit{DiConv3D} \left( \textit{ChConv3D}(X) \right) \right),
\label{eq:atten}
\end{equation}
where $X$ is the input tensor and $\textit{ChConv3D}$ is a depth-wise convolution operating on a single channel. $\textit{DiConv3D}$ is a dilated depth-wise convolution to broaden the receptive field and to enable the extraction of long-range dependencies. The point-wise convolution $\textit{Conv3D}_{1\times 1}$ is applied to aggregate the information across the channels. The final activations are obtained as follows:
\begin{equation}
\textit{Attention Value} = \textit{Atts} \odot X,
\label{eq:atten_res}
\end{equation}
where $\odot$ is the element-wise product.
The remaining components of the transformer block follow the conventional structure of typical transformers.

The decoder in our model aims to restore the spatial resolution of the input using a sequence of blocks (green circles in Figure \ref{fig:ulkanet}). Each decoder block consists of a chain of three convolution modules followed by an upsampling operation. The convolution modules are responsible for volumetric convolution operation. They consist of a Conv3D layer and a batch normalization layer, followed by a LeakyReLU activation function. The upsampling operation scales the resolution by a factor of two. As we stated earlier, a second larger illustration of our architecture that shows the inner modules can be found in the appendix section \ref{appendix_ulkanet}.

For each individual block in the encoder, the decoder has one corresponding block. There is also an additional decoder block in the bottleneck layer, as shown in Figure \ref{fig:ulkanet}. The input to each decoder block is supplied by the block in the previous layer and also the corresponding encoder block through a skip connection. In order to enhance the reconstruction of input, we use the skip connections to facilitate the transfer of high-level features \cite{YosinskiCBL14} to the layers that are responsible for the reconstruction task.

\begin{figure}[ht!]
  \centering
  \includegraphics[scale=0.43]{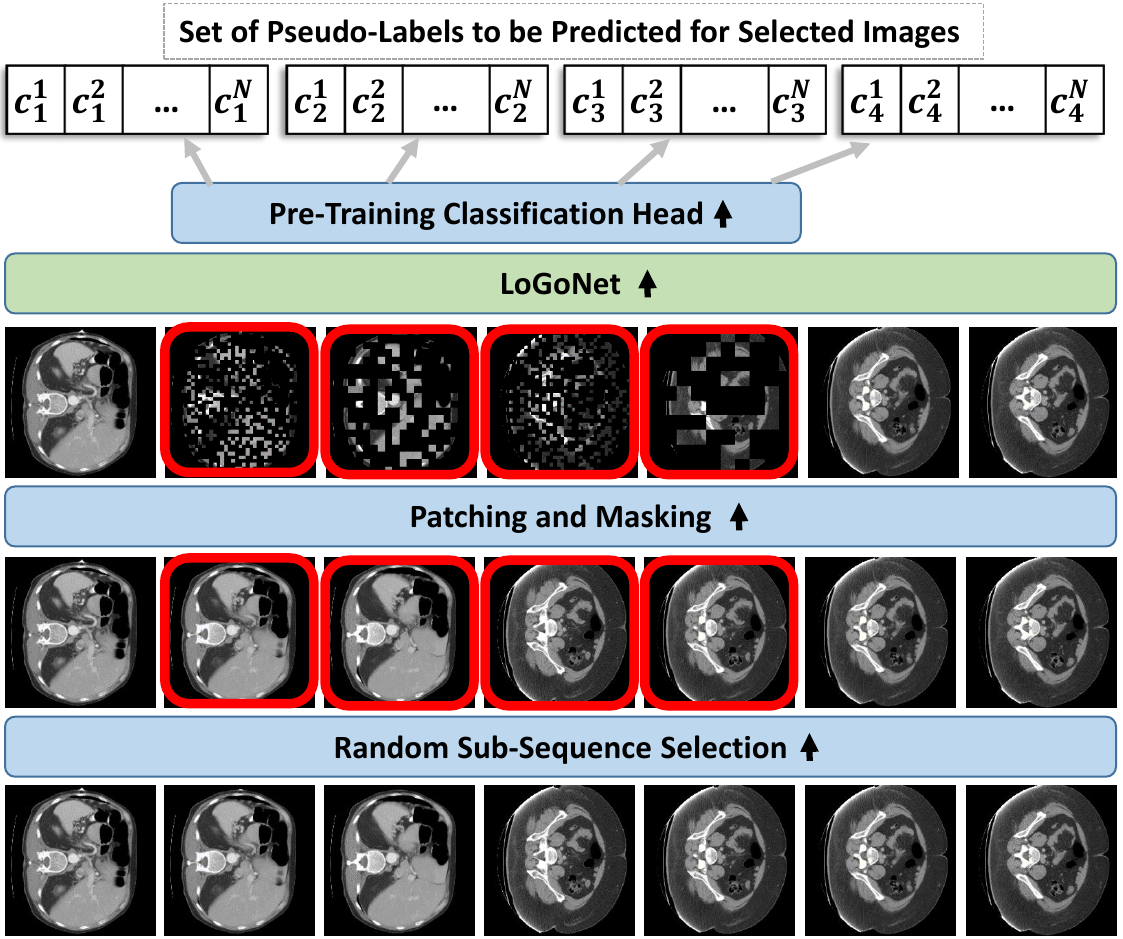}
  \caption{Illustration of our pre-training pipeline. We begin by randomly selecting a set of $m$ sequential images (here $m$ is four), on which we apply patching and masking. Then LoGoNet is used to predict the set of pseudo-labels that we generated for each distorted image (see Section \ref{sub-sec:pretrain} for details). During the pre-training stage, a classification head (a feed-forward network) is used on top of the model for prediction. This head is replaced with a convolution head (see Figure \ref{fig:overview}) for fine-tuning on the segmentation task with labeled data.}
  \label{fig:masked_LoGo}
\end{figure}

\subsection{Dual Encoding Strategy: An Alternative to Increasing Model Capacity} \label{sub-sec:dual-enc}

One of the difficulties in medical image segmentation is the presence of organs that have complex shapes. For instance, the human gallbladder and spleen have an elongated structure. Hence, to achieve satisfactory performance in the segmentation task, the model should be able to detect and extract relevant features in multiple regions of the input images, heavily relying on global features. On the other hand, this organ has irregular corners. This characteristic requires the model to be able to detect local features in multiple regions of the input. While increasing the model capacity by adding more layers, and also composing larger training sets, will potentially enable the model to automatically learn these regularities, this will likely increase costs during both the deployment and development stages.

To reduce the burden of automatic feature mining and, consequently, to reduce the costs, we propose to impose an inductive bias \cite{ml-book} on the feature extraction process. We propose to have two feature extractors in parallel, one focusing on the global scale and another one focusing on the local scale--as shown in Figure~\ref{fig:overview}. The global module is able to extract long-range dependencies due to access to the original data cube. On the other hand, the local module focuses on short-range dependencies. This is accomplished by partitioning the input cube into smaller ones, allowing for a more focused analysis and resulting in finer-grained features.

To implement our idea, we use one instantiation of ULKANet in the global module, and a sequence of $N$ instantiations of ULKANet in the local module. In the analysis section, we show that while using only one ULKANet can reduce the model size and speed up inference, it will also significantly deteriorate prediction accuracy. Additionally, we show that alternative strategies, used in comparable models, are either slower or achieve lower prediction accuracy. To prepare the data for the local module, the input 3D image is split into $N$ smaller cubes of size $B \times B \times B$. Given an image of size $S \times H \times W$, the value of $B$ is obtained by $B=\sqrt[3]{\frac{S \times H \times W}{N}}$.

To reconstruct the input data cube, the outputs of the local module are concatenated, as shown in Figure~\ref{fig:overview}. In order to aggregate the outputs of the global and local modules, we use an element-wise summation operator. The resulting tensor is expected to represent both global and local range dependencies.

\subsection{Pre-Training Method: Exploiting Unlabeled Images} \label{sub-sec:pretrain}

Before fine-tuning our model on labeled data, we utilize a multi-task pre-training technique to relocate the model weights to a favorable state. This self-supervised approach allows the model to learn general information from 3D medical images, without the necessity of ground-truth labels.

Pre-training of our model is done in three stages. First, we methodically mask certain regions of the input images. In this stage, the goal is to capture long-range and short-range feature dependencies. Second, we generate pseudo-labels for the masked images. The model later learns to generalize to unseen cases by predicting the pseudo-labels of the masked data. Finally, the masked images, along with their pseudo-labels, are used to pre-train the model. Below we explain each step.

\subsubsection{Masking Algorithm} 

In 3D imaging, objects are depicted across multiple 2D surfaces. Therefore, we argue that an effective masking strategy should step beyond 2D inputs.

In order to help the model explore not only the dependencies between pixels in 2D images but also the connections among pixels that form 3D masses, we propose an algorithm to mask chains of patches in an image sequence.\footnote{Note that in speech processing, where data is naturally sequential, applying this technique seems to be the default method \cite{hsu2021hubert}. However, to our knowledge, we are the first to propose this technique in the computer vision domain.} We begin by randomly selecting an image from the set of unlabeled data, with probability $\phi_1$ for selecting an individual image. Along the selected image, we also retrieve the $m - 1$ preceding images in the same sequence. Then, we apply a masking technique to the images in the chain. Various masking techniques can be used in this stage \cite{shi2022adversarial, li2021mst}; we employ the method introduced by \citet{xie2022simmim}. Therefore, for each image in the chain, we randomly select a patch size $P$, and partition it into $\frac{H \times W}{(P_j)^2}$ patches, where $H$ and $W$ are the height and width of the image. Finally, with the probability $\phi_2$ we mask out each patch of the image. Appendix section \ref{app:pre-train-details} discusses more details about the masking algorithm, and how to tune the hyperparameters.

In contrast to algorithms such as SimMIM \cite{xie2022simmim}, our proposed approach distinguishes itself by selecting a sequence of images and subsequently applying masking to that sequence. This method facilitates the encoder in gathering information by focusing on the interdependence of voxels within the sequence of images. Notably, our algorithm operates independently of the specific model structure, diverging from approaches seen in studies by \citet{kakogeorgiou2022hide}, \citet{he2022masked}, and \citet{zhou2021ibot}, all of which exhibit a reliance on model structure. Furthermore, our approach is compatible with Vision Transformer (ViT)-based and CNN-Based models.

\subsubsection{Pseudo-Label Generation}

Our pseudo-label generation algorithm assigns labels to all the images in the unlabeled set. Later in the pre-training pipeline, our model is asked to predict the pseudo-labels of the masked out images in each sequence. The information conveyed by the distorted images is insufficient for label prediction. Therefore, the model must explore the associations between pixels across multiple 2D images in the sequence to correctly predict the pseudo-labels of the target images. In the analysis section, we empirically show that this exploration task helps the model to learn the properties of the domain and to generalize better.

A clustering algorithm is employed for the pseudo-label generation. For simplicity, we use the k-means clustering method, although other types of clustering methods, such as hierarchical or spectral methods, can be utilized. Given a random number $k$ as the predefined number of clusters, we train a k-means clusterer on a random subset (e.g. $10\%$ in our experiments) of the unlabeled data. Then we use the clusterer to label the entire unlabeled set. Note that masking is not applied in any of these stages, and the clusterer has access to the unmasked images. The obtained labels are used as pseudo-labels to pre-train the model by predicting the corresponding labels for every masked image.

The k-means clusterer is able to use all the properties of the images to form the clusters. For instance, a cluster may constitute images that illustrate elongated organs, while another cluster may constitute images that depict organs that have particular corners. During pre-training, the model is asked to recover the pseudo-labels of a sequence of images that are distorted by masking. In order to predict their correct labels, the model must discover the associations between neighboring pixels. This pretext task enables the model to learn long-range and short-range spatial dependencies effectively.

Assuming that a clustering method exploits a finite set of characteristics in data to form the clusters, our model needs to learn these characteristics to correctly assign each image to the associated clusters. We conjecture that having $N$ different clusterers labeling the data and then using our model to simultaneously predict these multiple labels can further help the model gain broader knowledge from the data. From a different perspective, we can assume that recovering the characteristics of each clusterer is a separate pre-training task, and then, concurrently recovering the characteristics of multiple clusterers is a multi-task training. The efficacy of multi-tasking is well-documented in the machine learning literature \cite{multi-task}. Figure \ref{fig:masked_LoGo} shows our pre-training pipeline. In this figure, $N$ denotes the total number of clusterers, and $c_{i}^{j}$ denotes the pseudo-label generated by $j$-th clusterer for the $i$-th masked image in the sequence.

\subsubsection{Pre-Training Loss Function}

To pre-train our model, we use a cumulative negative log-likelihood function on the model predictions for the masked images as follows:
\begin{equation}
\mathcal{L} = - \sum_{i=1}^{N} \sum_{j=1}^{S} \log(p^i(e|x_j)),
\label{eq:loss}
\end{equation}
where $N$ is the number of clusterers, $S$ is the number of masked images that can be calculated by $S=M \times Q$, where $M$ is the length of image sequence for masking, and $Q$ is the number of concurrent masked sequences, if present. $x_1,x_2,x_3,...,x_S$ are masked images, and $p^i(e|x_j)$ is the probability that the $j$-th masked image in the sequence (i.e., $x_j$) is correctly assigned to the pseudo-label $e$ generated by the $i$-th clusterer. The value of $p^i(e|x_j)$ is calculated by a softmax function on top of the pre-training classification head, which is a simple feed-forward network.\footnote{Replacing the pre-training head with a finetuning head is an established practice in the self-supervision literature \cite{DevlinCLT19}.} Therefore, given a clusterer, we have:
\begin{equation}
p^{\parm}(e|x) = \frac{exp({f_e(x) / \tau
})}{\sum_{s=1}^{K} exp({f_s(x) / \tau
})},
\label{eq:softmax}
\end{equation}
where $exp(\parm)$ is the exponential function, $K$ is the number of clusters generated by the clusterer, $e$ is the cluster that the input image $x$ belongs to, and $f_s(x)$ is the $s$-th logit of the pre-training classification head. The hyper-parameter $\tau$ is called the softmax temperature. The value of $\tau$ determines the strength of the gradients backpropagated through the network. Lower temperature values increase the magnitude of gradients \cite{hinton2015distilling}. This, in turn, reduces the standard deviation of output probabilities--also known as sharpening the posterior probabilities.

Our loss function (Equation \ref{eq:loss}), iterates over all the predictions that our model makes during the pre-training stage and penalizes for the errors. As we discussed earlier, our pretraining framework enables LoGoNet to become familiar with the properties of the domain to generalize better by exploiting unlabeled data. We empirically support this argument in our analysis section. Additional experiments can be found in appendix section \ref{app:pre-train-details}.

\begin{table*}
\centering
\begin{tabular}{c|ccccc}
\hline
\textbf{Models}    & \textbf{SegResNetVAE} & \textbf{SwinUNETR}      & \textbf{UNETR}           & \textbf{UNet++}       & \textbf{nnUNet} $\rightarrow$       \\ \hline
\textbf{FLOPs (G)} & 15.50                                  & 329.84                                   & 264.59                                    & 4229.20                                & 1250.65                                \\
\textbf{\# Param}  & 3.9 M                                  & 62.2 M                                   & 101.7 M                                   & 84.6 M                                 & 30.7 M                                 \\ \hline
$\rightarrow$ \textbf{Models}    & \textbf{DiNTS Search} & \textbf{DiNTS Instance} & \textbf{Attention U-Net} & \multicolumn{2}{c}{\textit{\textbf{LoGoNet}}} \\ \hline
\textbf{FLOPs (G)} & 743.88                                 & 743.88                                   & 7984.21                                   & \multicolumn{2}{c}{246.96}    
                        \\
\textbf{\# Param}  & 74.1 M                                 & 74.1 M                                   & 64.1 M                                    & \multicolumn{2}{c}{67.5 M}                              
\\ \hline

\end{tabular}
\caption{Comparison between our model and the baselines in terms of inference speed (in floating-point operations per second) and the number of trainable parameters in the BTCV dataset. Due to the size of the images, the results are identical across the BTCV and MSD datasets. See appendix section \ref{app:imp_det} for more experiments on resource consumption.}
  \label{tab:BTCV_baselines}
\end{table*}

\section{Experimental Setup} \label{sec:experimental_setup}

In this section, we briefly describe the datasets used in the experiments, provide a list of baseline models we compare to, and also provide an overview of our setup.

\noindent\textbf{Datasets.} We use two widely used standard datasets. As the first dataset, we use the BTCV dataset\footnote{Available at \url{https://www.synapse.org/\#!Synapse:syn3193805/wiki/217789}} introduced by \citet{gibson2018automatic}. This dataset contains 13 segmentation tasks, and each task has 40 data points obtained via abdominal CT scans. As the second dataset, we use the MSD dataset\footnote{Available at \url{http://medicaldecathlon.com/}} introduced by \citet{simpson2019large}. This dataset contains a variety of tasks obtained via magnetic resonance imaging (MRI), computed tomography (CT), and positron emission tomography (PET). We use six different tasks from this dataset that contain a total of 900 examples. The MSD dataset contains 6 tasks, of which 4 are cancer or tumor detection (anomaly detection), e.g., colon cancer. As the unlabeled data, we use the meta-dataset collected by \citet{tang2022self}, which consists of 4,500 examples. The images in this dataset are not annotated, and are 3D scans covering a variety of organs.

\noindent\textbf{Baselines.} We compare LoGoNet to a suite of baseline models, including those that use Visual Transformers or Convolutional Neural Networks. We compare to nnUNet \cite{isensee2018nnu}, Attention U-Net \cite{oktay2018attention}, SegResNetVAE \cite{myronenko20193d}, UNet++ \cite{zhou2019unet++}, DiNTS (two variations of Search and Instance) \cite{he2021dints}, SwinUNETR (feature size 48) \cite{hatamizadeh2021swin}, and UNETR (feature size 32) \cite{hatamizadeh2022unetr}.

\noindent\textbf{Setup.} We follow standard practices to carry out the experiments. We use the Dice metric, a common metric for the image segmentation task, to report the performance results. We conduct the experiments in each dataset task separately and report the average results for five runs in the BTCV dataset and two runs in the MSD dataset. Detailed information about hyperparameter tuning, configurations, and implementation is reported in the appendix section~\ref{app:imp_det}.

Our default LoGoNet and ULKANet models have four encoder blocks with 3, 4, 6, and 3 transformer modules in each block, respectively. The dimensions of the embedding vectors in these models are 64, 128, 256, and 512, respectively.

\subsection{Pre-Training Details}
\label{app:pre-train-details}

We used the scikit-learn implementation\footnote{Available at: https://scikit-learn.org/stable/} of the Mini Batch KMeans algorithm as the clusterers in our pre-training pipeline. The outcomes of vanilla K-means clustering are unstable, and this can make the reproducibility challenging. To address this problem, we used K-means++ (implemented in Mini Batch KMeans). K-means++ addresses this issue directly through its enhanced seeding process. It improves the stability and reproducibility of clustering results by systematically selecting initial centers to reduce the variability caused by random initialization in standard k-means.

\begin{table}[ht]
\begin{tabular}{ll}
\hline
\textbf{Configuration}                            & \textbf{Value}                                                                                                               \\ \hline
Optimizer                                & $AdamW$                                                                                                               \\
Epochs                                   & $100$                                                                                                                 \\
Batch Size per GPU                       & $1$                                                                                                                   \\
Number of GPUs                           & $16$                                                                                                                  \\
Weight decay                             & $1e-5$                                                                                                                \\
Optimizer momentum                       &  $\beta_1$, $\beta_2$ = $0.9$, $0.999$                                                       \\
Peak learning rate                       & $1e-4$                                                                                                                \\
Learning rate schedule                   & $CosineAnnealingLR$                                                                                       \\
Warmup epochs                            & 10                                                                                                                  \\
Dropout                                  & 0                                                                                                                   \\
Rand Spatial Crop Samples Data           & $96 \times 96 \times 96 $                                                             \\
MONAI Transforms: ScaleIntensityRanged & \begin{tabular}[c]{@{}l@{}}$a\_min$ = -1000 \\ $a\_max$ = 1000\\ $b\_min$ = 0\\ $b\_max$ = 1\\ Clip = True\end{tabular} \\
$\tau$ & 0.1 \\

 $\phi_1$                 & $0.1$                                                                                                                \\
$\phi_2$                 & $0.7$                                                                                                                \\
M (Size of masked sequence)              & $5$                                                                                                                   \\
$P_j$ (Size of Patches)                & $1, 2, 4, 8, 16, 32, 96$                                                                                              \\ \hline
\end{tabular}

  \caption{Pre-Training settings for our proposed approach}
  \label{tab:pre-train-info}
\end{table}

During the training phase of the k-means models, we adopted a transformation process that converted the input image from a $Channel \times X \times Y \times Z$ format to a vector representation of dimensions $Z \times T$, where $T$ is equivalent to $Channel \times X \times Y$. This transformation enabled us to generate a label for each cluster per image slice, resulting in a sequence of labels for a sequence of images. Subsequently, the model underwent $350$ iterations of training, with each iteration utilizing a randomly selected $10\%$ subset of the unlabeled data. To introduce diversity and enhance robustness, we employed a stochastic approach in determining the value of $K$, randomly sampling from a range spanning $80$ to $500$.

The information pertaining to pre-training is reported in Table \ref{tab:pre-train-info}. To pre-train the model, we leveraged the $AdamW$ optimizer, and set the hyperparameters $\phi_1$ to $0.1$ and $\phi_2$ to $0.7$. Additionally, the sequence of distorted images, denoted as $M$, was set to $5$.

Our observations reveal that augmenting both the values of $M$ (length of sequenced mask images) and $\phi_1$ (rate of sampled images) results in an increased rate of masked images. However, this heightened rate poses challenges to our model during the pre-training, and enables it to exploit dependencies between successive slices for effectively capturing information related to missing voxels. This delicate interplay between hyperparameters emphasizes the necessity of finding an optimal balance to enhance model performance, as an excessive increase in masked images may impede the model's ability to leverage contextual dependencies within the data.

Furthermore, we introduced randomness in the selection of patch sizes, choosing from the set $(1, 2, 4, 8, 16, 32, 96)$. Our pre-training approach involves the incorporation of a classification head designed to adapt the model output to align with the requirements of our pseudo-labeling. Figure \ref{fig:masked_LoGo} shows the structure of our proposed pre-training. The structure of the classification head can be found in Algorithm \ref{alg:pre-train-head}.

\begin{algorithm}
\caption{Pseudo Code of Pre-Training Classification Head}
\label{alg:pre-train-head}
\begin{algorithmic}[1]
\Procedure{PreHead}{$X$, $input\_dim$, $x\_dim$, $y\_dim$, $z\_dim$, $cluster\_num$, $class\_size$} 
\Comment{Input: $X$ is the input tensor.}
    \State $X \gets Conv3d(X, input\_dim, cluster\_num)$
    \State $X \gets BatchNorm3d(X, cluster\_num).GELU(X)$  
    \State $X \gets Conv3d(X, cluster\_num, cluster\_num)$ 
    \State $X \gets BatchNorm3d(X, cluster\_num).GELU(X)$
    \State $X \gets X.permute(0, 3, 2, 1, 4)$
    \newline
    \State $X \gets Conv3d(X, y\_dim, class\_size)$
    \State $X \gets BatchNorm3d(X, class\_size).GELU(X)$  
    \State $X \gets Conv3d(X, class\_size, class\_size)$ 
    \State $X \gets BatchNorm3d(X, class\_size).GELU(X)$ 
    \State $X \gets X.permute(0, 2, 1, 3, 4)$
    \newline
    \State $X \gets Conv3d(X, x\_dim, x\_dim // 16)$
    \State $X \gets BatchNorm3d(X, x\_dim // 16).GELU(X)$  
    \State $X \gets Conv3d(X, x\_dim // 16, 1)$ 
    \State $X \gets BatchNorm3d(X, 1).GELU(X)$
    \State $X \gets X.permute(0, 4, 3, 2, 1)$
    \newline
    \State $X \gets Conv3d(X, z\_dim, z\_dim)$
    \State $X \gets BatchNorm3d(X, z\_dim).GELU(X)$
    \State $X \gets Conv3d(X, z\_dim, z\_dim)$
    \State $X \gets ReLU(X).squeeze()$

    \State \textbf{Return} X
\EndProcedure
\end{algorithmic}
\end{algorithm}

\subsection{Fine-Tuning Details}
\label{app:conf-setup}

Table \ref{tab:train-info} provides a comprehensive overview of the specifics pertaining to our training or fine-tuning procedures.

\begin{table}[]
\begin{tabular}{ll}
\hline
\textbf{Configuration}                          & \textbf{BTCV}                                                                                                                                                                                                            \\ \hline
Optimizer                              & AdamW                                                                                                                                                                                                    \\
Epochs                                 & 5000                                                                                                                                                                                                       \\
Batch Size per GPU                     & 2                                                                                                                                                                                                              \\
Number of GPUs                         & $16$                                                                                                           \\
Weight decay                           & $1e-5$                                                                                                                                                                                                      \\
Optimizer momentum                     & $\beta_1$, $\beta_2$ = $0.9$, $0.999$                                                                                                                                             \\
Peak learning rate                     & $1e-4$                                                                                                                                                                                                      \\
Learning rate schedule                 & CosineAnnealingLR                                                                                                                                                                            \\
Warmup epochs                          & $100$                                                                                                                                                                                                         \\
Dropout                                & 0                                                                                                                                                                                                        \\
Rand Spatial Crop Samples Data         & $96 \times 96 \times 96$                                                                                                                                                                                                                   \\ \hline
\end{tabular}

  \caption{Training and fine-tune settings for all proposed and baseline models}
  \label{tab:train-info}
\end{table}

Our experimental setup involved 16 GPUs, each one having 35GB of memory. The models were trained with $AdamW$ optimizer. As we stated earlier, our goal was to design a fast model at inference time, as such models are used countless times in the deployment environment and can have a large carbon footprint. As shown in Table \ref{tab:BTCV_baselines}, our model's required resources in terms of FLOPs are one of the best, lower than 7 of the baselines with which we compared our model to. Our model's total number of parameters is also lower than many baselines while achieving higher performance. We used multiple GPUs for training and data parallelism to decrease the time required for training. However, our model is small enough to be trained on a single GPU with 14 GB capacity, so researchers whose access to GPUs is limited can still train and validate our model.

To adhere to established standards and foster equitable comparisons, we employed a comprehensive array of augmentation techniques to augment data variability. It's noteworthy that these augmentations were uniformly applied to all models, encompassing both our proposed models and the baseline models. This meticulous approach ensures a fair and unbiased comparative analysis. For the implementation of our models and the baseline models, we leveraged the $MONAI$ framework,\footnote{Available at: https://monai.io/} which provided a robust and versatile foundation for our experimentation. This framework facilitated the seamless integration of existing public implementations.

In the course of each iteration, we implemented a randomized cropping strategy, extracting two images for each case during the training phase. This deliberate approach was employed with the intent of diversifying the training dataset for each input case within every epoch, thereby enhancing the overall richness of the training process.

\section{Results} \label{sec:results}

\subsection{Main Results}

\begingroup 
\setlength{\tabcolsep}{4pt} 

\begin{table*}[]
  \centering

\begin{tabular}{c|ccccccccccccc|c}
\hline
Models                 & Spl           & RKid          & Lkid          & Gall          & Eso           & Liv           & Sto           & Aor           & IVC           & Veins         & Pan           & Rad           & Lad           & AVG           \\ \hline
UNETR                  & .912          & .940          & .938          & .693          & .690          & .954          & .754          & .891          & .830          & .703          & .734          & .660          & .577          & .790          \\
SegResNetVAE           & .941          & .938          & .933          & .670          & .718          & .955          & .745          & .892          & .848          & .695          & .783          & .633          & .528          & .791          \\
nnUNet                 & .859          & .944          & .924          & .796          & .755          & .960          & .781          & .894          & .849          & .756          & .776          & .675          & .663          & .818          \\
Attention U-Net        & .955          & .936          & .930          & .735          & .739          & .964          & .770          & .898          & .852          & .753          & .763          & .695          & .688          & .821          \\
DiNTS Instance         & .935          & .942          & .938          & .770          & .769          & .962          & .743          & .909          & .857          & .759          & .782          & .641          & .691          & .823          \\
UNet++                 & .934          & .931          & .925          & .810          & .715          & .961          & .786          & .900          & .846          & .747          & {\ul .829}    & .685          & .679          & .827          \\
SwinUNETR              & .952          & {\ul .947}    & {\ul.945} & .790          & .770          & .963          & .755          & .901          & .850          & {\ul .771}    & .760          & .702          & .659          & .828          \\
DiNTS Search           & .937          & .934          & .930          & .788          & .770          & .960          & .774          & .904          & {\ul .866}    & .751          & .813          & .670          & {\ul .711}    & .831          \\ \hline
\textbf{LoGoNet}       & {\ul .958}    & \textbf{.949} & \textbf{.947} & {\ul .818}    & {\ul .786}    & {\ul .969}    & {\ul .880}    & {\ul .912}    & .865          & .769          & .821          & {\ul .726}    & .698          & {\ul .854}    \\
\textbf{LoGoNet + PRE} & \textbf{.961} & {\ul .947}    & .944          & \textbf{.866} & \textbf{.845} & \textbf{.970} & \textbf{.898} & \textbf{.936} & \textbf{.885} & \textbf{.791} & \textbf{.838} & \textbf{.738} & \textbf{.757} & \textbf{.875} \\ \hline
\end{tabular}

  \caption{Performance of our model (in terms of Dice metric) compared to the baseline models in BTCV dataset. All experiments were conducted using identical data splits, computing resources, and testing conditions to ensure a fair comparison. Additionally, to ensure faithfulness to the original implementation of the baseline methods, we used their publicly available implementations available at MONAI network repository. Spl: Spleen, RKid: Right Kidney, LKid: Left Kidney, Gall: Gallbladder, Eso: Esophagus, Liv: Liver, Sto: Stomach, Aor: Aorta, IVC: Inferior Vena Cava, Veins: Portal and Splenic Venis, Pan: Pancreas, Rad: Right Adrenal Glands, Lad: Left Adrenal Glands.}
  \label{tab:BTCV_res}
\end{table*}

\endgroup 

Table \ref{tab:BTCV_baselines} compares our model to the baseline methods in terms of inference time (FLOPs) and the number of trainable parameters in the BTCV dataset. We see that our model has the lowest inference time after SegResNetVAE. Tables \ref{tab:BTCV_res} and \ref{tab:MSD_res} compare the accuracy of our method to the baselines. We observe that the performance of SegResNetVAE is significantly lower than that of ours. Taking into account both the inference speed and the prediction accuracy, our model seamlessly ranks first among all the models. See Appendix \ref{sec:com_results} for a report on the standard deviation and statistical significance of the results.

Table \ref{tab:BTCV_baselines} shows that our model is considered an average-sized network. One noteworthy observation is that in some cases, e.g., nnUNet or DiNTS Instance, even though the number of trainable parameters is on a par or smaller than ours, their inference speed is substantially slower. Tables \ref{tab:BTCV_res} and \ref{tab:MSD_res} show that our model exhibits superior performance compared to the baseline methods. Specifically, when evaluating our proposed model without pre-training, it outperforms the baselines across $13$ out of $19$ tasks. Furthermore, incorporating our pre-training strategy into LoGoNet enhances its performance even further, surpassing the baselines in $18$ out of $19$ tasks. These findings underscore the effectiveness and versatility of our approach in tackling a diverse range of tasks with notable efficacy.

\begingroup 
\setlength{\tabcolsep}{2pt} 

\begin{table}[ht]
\centering

\begin{tabular}{c|cccccc|c}
\hline
\textbf{Models}                 & \textbf{Col}           & \textbf{Spl}           & \textbf{Hep}           & \textbf{Pan}           & \textbf{Lun}           & \textbf{Car}           & \textbf{AVG}           \\ \hline
UNETR                  & .677          & .969          & .715          & .699          & .730          & .953          & .790          \\
SegResNetVAE           & .742          & .968          & .745          & .740          & .765          & .951          & .818          \\
nnUNet                 & .736          & .977          & .742          & .742          & {\ul .816}    & .958          & .829          \\
Attention U-Net        & -             & -             & -             & -             & -             & -             & -             \\
DiNTS Instance         & .768          & {\ul .979}    & .731          & .742          & .790          & \textbf{.963} & .829          \\
UNet++                 & .553          & .975          & .752          & .760          & .753          & {\ul .961}    & .792          \\
SwinUNETR              & .695          & .967          & .737          & .738          & .763          & .957          & .810          \\
DiNTS Search           & .776          & \textbf{.980} & .749          & .749          & .768          & .960          & .830          \\

\hline
\textbf{LoGoNet}       & {\ul .786}    & \textbf{.980} & {\ul .757}    & {\ul .798}    & .802          & .951          & {\ul .846}    \\
\textbf{LoGoNet + PRE} & \textbf{.801} & \textbf{.980} & \textbf{.779} & \textbf{.833} & \textbf{.828} & .958          & \textbf{.863} \\ \hline
\end{tabular}
  \caption{Performance of our model (in terms of Dice metric) compared to the baselines in MSD dataset. The baseline model ``Attention U-Net'' was not runnable on regular chipsets which each has 35 Gigabyte of memory in MSD dataset. Col: Colon Cancer Primaries, Spl: Spleen, Hep: Hepatic vessels and tumor, Pan: Pancreas Tumour,  Lun: Lung Tumours, Car: Cardiac.}
  \label{tab:MSD_res}
\end{table}

\endgroup 

In the BTCV dataset, LoGoNet outperforms the top three baseline models on average by $2.7\%$, $3.0\%$, and $3.2\%$, respectively. Regarding the inference time, our model outperforms the top three models by $17.6\%$, $14.8\%$, and $118.2\%$, respectively.

\subsection{Analysis}

In this section, we demonstrate the properties of our model from multiple aspects. Specifically, we report a qualitative comparison between our model and the best baseline model, evaluate our strategy for extracting local and global features, evaluate our pre-training approach, show the impact of model size on performance, analyze the hyper-parameter sensitivity of our model, and finally, report an ablation study on the steps in our pre-training method. The experiments in this section are carried out in the BTCV dataset unless stated otherwise.




\begin{figure}[ht]
\small
  \centering
  \begin{tabular}{c|cccccc}
   \hline
    {Ground Truth} & \textbf{LoGoNet } & DiNTS Search \\
    \hline
    \includegraphics[width=2.25cm]{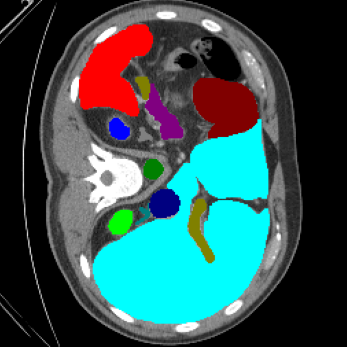} &
    \includegraphics[width=2.25cm]{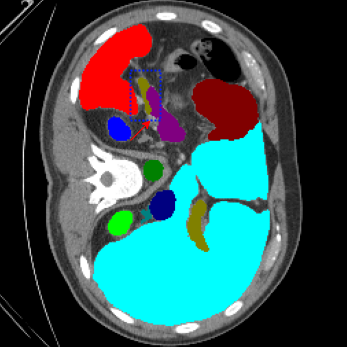} &
    \includegraphics[width=2.25cm]{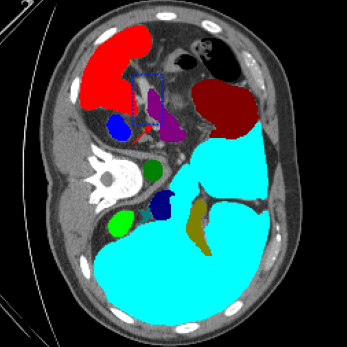}\\

    \includegraphics[width=2.25cm]{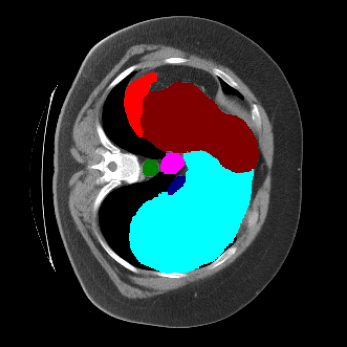} &
    \includegraphics[width=2.25cm]{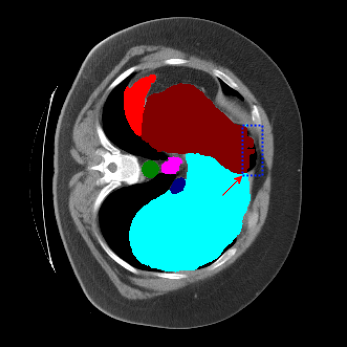} &
    \includegraphics[width=2.25cm]{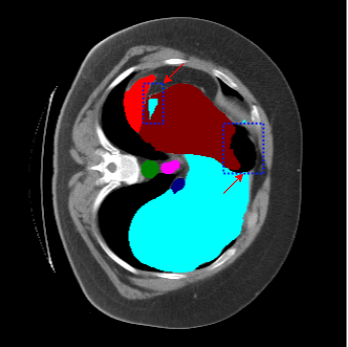}\\

    \includegraphics[width=2.25cm]{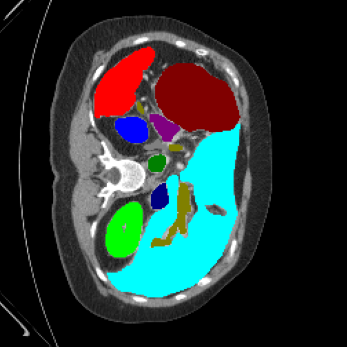} &
    \includegraphics[width=2.25cm]{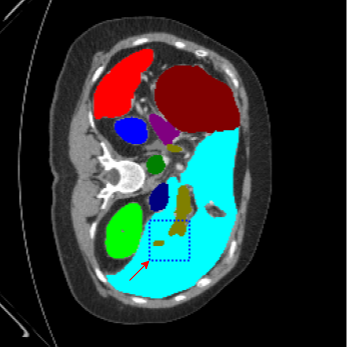} &
    \includegraphics[width=2.25cm]{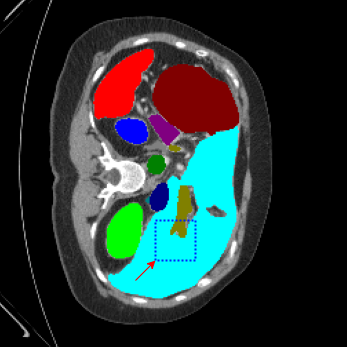}\\
    \hline
  \end{tabular}
  \caption{Output of LoGoNet compared to the best performing baseline model in BTCV dataset, i.e., DiNTS Search. We see that our model tangibly outperforms the mentioned model in detecting organ boundaries.}
  \label{fig:BTCV_res}
\end{figure}

We begin by qualitatively inspecting our model. Figure \ref{fig:BTCV_res} compares the output of LoGoNet to the best-performing baseline model in the BTCV dataset, i.e., DiNTS Search. We see that our model particularly excels in segmenting organ boundaries. This can be attributed to our effective strategy for extracting local-range dependencies, which plays a crucial role in extracting details from input data. Our model's adeptness in capturing long-range dependencies allows it to grasp contextual information that extends over significant distances within the data. Simultaneously, its proficiency in handling short-range dependencies ensures precision in capturing localized patterns.

\begin{table}[ht]
\begin{tabular}{c|cccc|c}
\hline
\textbf{Models}  & \textbf{Gall} & \textbf{Eso}  & \textbf{Veins} & \textbf{Lad}  & \textbf{AVG}  \\ \hline
\textbf{ULKANet} & .761          & .782          & .690           & .684          & .824          \\
\textbf{LoGoNet} & \textbf{.818} & \textbf{.786} & \textbf{.769}  & \textbf{.698} & \textbf{.854} \\ \hline
\end{tabular}

  \caption{The efficacy of our parallel strategy for extracting local and global features, i.e., the comparison between our method (LoGoNet) and an alternative method that relies on a single feature extractor (ULKANet).}
  \label{tab:logo_or_not}
\end{table}

To further quantitatively support our strategy for extracting local and global features in parallel, in the next experiment, we report the performance of our model compared to the regular method for extracting features from medical images, which is relying on a single feature extractor. This translates into comparing LoGoNet to our feature extractor ULKANet. Table \ref{tab:logo_or_not} reports the results. We observe that our strategy enables our model to outperform the alternative method.

In the next experiment, we report the efficacy of our pre-training method. To carry out this experiment, we use the algorithm proposed in Section \ref{sub-sec:pretrain} to initialize the weights of our model, and then, we follow the regular fine-tuning steps. In Tables \ref{tab:BTCV_res} and \ref{tab:MSD_res} (the last rows), we report the results of this model for both datasets, indicated by postfix PRE. We see that the improvements achieved by pre-training are consistent across both datasets.

\begin{table}[ht]
\begin{tabular}{c|cccc|c}
\hline
\textbf{SSL Approach}                                    & \textbf{Gall} & \textbf{Eso}  & \textbf{Veins} & \textbf{Lad}  & \textbf{AVG}  \\ \hline
SimMIM \cite{xie2022simmim}             & {\ul .837}    & {\ul .829}    & {\ul .785}     & {\ul .733}    & {\ul .864}    \\
Rubik’s Cube   \cite{tao2020revisiting} & .815          & .820          & .780           & .725          & .859          \\
SimCLR     \cite{chen2020simple}        & .829          & .803          & .780           & .720          & .859          \\ \hline
\textbf{Our SSL Approach}                                & \textbf{.866} & \textbf{.845} & \textbf{.791}  & \textbf{.757} & \textbf{.875} \\ \hline
\end{tabular}

  \caption{Performance of our multi-task self-supervised pre-training method compared to the alternatives (number of clusters is N=80).}
  \label{tab:res-ssl}
\end{table}

In the next experiment, we compare the effectiveness of our self-supervised pre-training approach to the alternative methods. In particular we compare to SimMIM \cite{xie2022simmim}, Rubuk's Cube \cite{tao2020revisiting}, and SimCLR \cite{chen2020simple} strategies. Table \ref{tab:res-ssl} reports the result. The numbers are obtained by initializing LoGoNet. Notably, our proposed model exhibits superior performance in three out of four experiments, showcasing its effectiveness in a diverse set of tasks. The comparison in Table \ref{tab:res-ssl} highlights the competitive edge of our model.

An inherent advantage of our pre-training approach lies in its versatility, as it is designed to be compatible with both CNN and ViT-based models. This flexibility broadens the applicability of our approach, allowing it to seamlessly integrate with different architectural paradigms commonly used in computer vision tasks.

\begin{table}[ht]
\begin{tabular}{c|cccc|c}
\hline
\textbf{Models}          & \textbf{Gall} & \textbf{Eso}  & \textbf{Veins} & \textbf{Lad}  & \textbf{AVG}  \\ \hline
\textbf{LoGoNet}         & .818          & .786          & .769           & .698          & .854          \\
\textbf{LoGoNet L}       & .847          & .781          & .768           & .710          & .855          \\ \hline
\textbf{LoGoNet + PRE}   & {\ul .866}    & {\ul .845}    & {\ul .791}     & {\ul .757}    & {\ul .875}    \\
\textbf{LoGoNet L + PRE} & \textbf{.921} & \textbf{.859} & \textbf{.805}  & \textbf{.784} & \textbf{.891} \\ \hline
\end{tabular}

  \caption{Performance of LoGoNet compared to LoGoNet L (Number of clusters N=80, L stands for the large model variant).}
  \label{tab:model_size_pretrain}
\end{table}

To understand the impact of model size on the prediction accuracy, we report the performance of our default model compared to a larger variant. Our larger variant uses four encoder blocks with 3, 3, 24, and 3 transformer modules, respectively. The dimensions of the embedding vectors in this model are 96, 192, 384, and 768, respectively. Table \ref{tab:model_size_pretrain} reports the results. Upon increasing the dimensions of our model, we observed an improvement in results, though it fell short of our initial expectations. We attribute this to the limited number of labeled data available. However, upon integrating our pre-training methodology into our standard and larger variants of LoGoNet, we noted a significant enhancement in performance, particularly noticeable in the larger LoGoNet.

\begin{table}[ht]
\centering
\begin{tabular}{c|cc|cc|cc}
\hline
\multirow{2}{*}{\textbf{Model}} & \multicolumn{2}{c|}{\textbf{N = 1}} & \multicolumn{2}{c|}{\textbf{N = 40}} & \multicolumn{2}{c}{\textbf{N = 80}} \\ \cline{2-7} 
                                & \textbf{Gall}     & \textbf{Eso}    & \textbf{Gall}     & \textbf{Eso}     & \textbf{Gall}     & \textbf{Eso}    \\ \hline
\textbf{LoGoNet + PRE}          & .830              & .819            & .843              & .860             & .866              & .845            \\ \hline
\end{tabular}

  \caption{Performance of our models at varying number of clusterers for pre-training. As the number of clusterers increases, the contribution of multi-tasking becomes more noticeable.}
  \label{tab:Ablation_res}
\end{table}

In Section \ref{sub-sec:pretrain}, we claimed that having multiple clusterers serves as a multi-task training approach. In order to demonstrate the benefit of having multiple clusterers, and also show the sensitivity of our model to the number of these learners in our algorithm, we report the results of our model with varying numbers of clusterers in Table \ref{tab:Ablation_res}. We see that as the number of clusterers increases, the performance improves. The results support our hypothesis regarding the ability of our model to extract broader knowledge from the unlabeled data in the presence of multi-tasking.

To refine our model using labeled data, we employed the DiceCELoss as the objective function during the fine-tuning or training process. The DiceCELoss function serves as a crucial metric, enabling us to strike a balance between the Dice coefficient and Cross-Entropy, optimizing the model's performance on the labeled dataset. The DiceCELoss is articulated by the following formulation: 
\begin{equation}
\small
DiceCELoss = w_{dl} \times DiceLoss + w_{cl} \times CELoss,
\label{eq:DiceCELoss}
\end{equation}
where
\begin{equation}
\small
DiceLoss = 1 - \frac{{2 \times \sum_{{i=1}}^{N} p_i \times t_i + \epsilon}}{{\sum_{{i=1}}^{N} p_i + \sum_{{i=1}}^{N} t_i + \epsilon}},
\label{eq:DiceLoss}
\end{equation}
and 
\begin{equation}
\small
CELoss = -\frac{1}{N} \sum_{{i=1}}^{N} t_i \times \log(p_i).
\label{eq:CELoss}
\end{equation}

Thus, our fine-tuning and training loss term is the weighted summation between the regular dice loss term and the cross entropy term. $p_i$ represents the predicted probability for the $i$-th class. $t_i$ represents the ground truth label for the $i$-th class. $N$ represents the number of classes. $\epsilon$ is a small constant (e.g., 1e-5) added to the denominator to avoid division by zero.

\begin{table}[ht!]
\begin{tabular}{c|cc|cc|cc}
\hline
\multirow{3}{*}{\textbf{LoGoNet}} & $w_{dl}$   &  $w_{cl}$   & $w_{dl}$   & $w_{cl}$   & $w_{dl}$  &  $w_{cl}$  \\ \cline{2-7} 
                         & 1.0           & 1.0           & 0.0           & 1.0           & 1.0           & 0.0          \\ \cline{2-7} 
                         & \multicolumn{2}{c|}{\textbf{.854}} & \multicolumn{2}{c|}{.841} & \multicolumn{2}{c}{{\ul.847}} \\ \hline
\end{tabular}
\caption{Performance outcomes with varied weights for DiceCELoss: The presented results represent the average across all 13 organs in the BTCV dataset using the LoGoNet model.}
  \label{tab:loss_diff}
\end{table}

Our experiments revealed that assigning equal weights to both CELoss and DiceLoss yields more favorable outcomes, surpassing the performance achieved with other weight ratios. The results of various weight configurations for losses are presented in Table \ref{tab:loss_diff}. By according equal significance to both Cross-Entropy Loss (CELoss) and Dice Loss, we strike a balance that enhances the model's ability to effectively capture diverse patterns in the data.

Finally, we report an ablation study on the effectiveness of our masking approach during the pre-training stage. In Section \ref{sub-sec:pretrain}, we argued that by distorting input images, the model must learn the properties of neighboring pixels in order to predict the correct labels. We then argued that this exploration task enables the model to faster learn the domain and to generalize better. The results reported in Table \ref{tab:Ablation_res_mask} supports our claim. We see that by incorporating the masking step, the performance noticeably improves signifying a better generalizablity of our method.


\begin{table}[ht]
\begin{tabular}{c|cc|cc|cc}
\hline
\multirow{2}{*}{\textbf{Model}} & \multicolumn{2}{c|}{\textbf{w/ M}} & \multicolumn{2}{c|}{\textbf{wo/ M}} & \multicolumn{2}{c}{\textbf{w/ M + wo/ M}} \\ \cline{2-7} 
                                 & \textbf{Gall}    & \textbf{Eso}    & \textbf{Gall}     & \textbf{Eso}    & \textbf{Gall}        & \textbf{Eso}       \\ \hline
\textbf{LoGoNet + PRE}           & .866             & .845            & .845              & .802            & .851                 & .820               \\ \hline
\end{tabular}

\caption{Ablation study on the effectiveness of our masking algorithm for 3D inputs. "w/ M" refers to pretraining with masking, and "wo/ M" refers to pretraining without masking. (BTCV Dataset)}
  \label{tab:Ablation_res_mask}
\end{table}





\subsection{Complexity analysis}

This section presents a comprehensive analysis of the computational complexity associated with our models, detailing the number of trainable parameters and the FLOPs. Please refer to Table \ref{tab:BTCV_baselines} for a summary of these metrics.

To make our computations more manageable, we simplify by excluding biases. Let's assume an input of size $C \times Z \times W \times H$, where $C$ represents the number of channels and $Z$, $W$, $H$ denote the spatial dimensions. From here, we derive the complexity expression. Specifically, with a kernel size of $K$ and a mean dilation rate of $d$, the complexity can be expressed as $O(((K/d)^2 \times C + (2 \times d - 1)^2 + C) \times C \times W \times H \times Z)$. This is how we arrive at our computational complexity.

It's worth noting that both $d$ and $K$ are constants in our system. This simplifies the complexity to $O(C^2 \times Z \times W \times H \times e)$, where $e$ represents a constant value. This single LKA block complexity is a direct result of these constants in our system.

Extending this analysis to encompass a network architecture with $T$ blocks in the encoder, each containing $L$ number of LKA blocks, the overall complexity becomes $O(C^2 \times Z \times W \times H \times T \times L \times e)$, encapsulating the computational demands of the entire system.

Furthermore, our complexity analysis provides valuable insights into the computational demands of our proposed models. By simplifying computations and excluding biases, we derive a comprehensive understanding of the system's scalability and efficiency. Notably, with each LKA block complexity being a direct consequence of constant parameters, the scalability of our system becomes evident. Extending this analysis to encompass the entire network architecture, comprising multiple blocks in the encoder, we obtain a holistic view of the computational complexity, highlighting its manageable nature even in large-scale implementations.

In summary, we demonstrated the efficacy of our model in two datasets across 19 segmentation tasks. We also compared our method to eight recent baseline models, including those that use Visual Transformers. Our results testify to the effectiveness of our novel feature extraction techniques. Our analysis shows that our pre-training method is successfully able to exploit unlabeled data to improve parameter initialization. We also showed that our method significantly speeds up inference time compared to the best-performing models.

Computer vision domain is a rapidly evolving research field. It seems unrealistic to expect long-term plans, specifically considering the rise of large pretrained vision models. However, with the existing challenges in the medical domain, this community will invest more in developing methods for mitigating the lack of large labeled sets. Therefore, in the next step, we plan to explore Domain Adaptation, which is one of the well-known methods for addressing this challenge.

\section{Conclusions}

In this paper, we proposed a fast and accurate approach for 3D medical image segmentation termed LoGoNet, which facilitates the augmentation of global and local feature dependencies. The localized mechanism in LoGoNet significantly improves segmentation, especially for small organ sections, while the incorporation of both global and local dependencies enhances the segmentation accuracy for elongated organs. We further proposed a pre-training method to exploit unlabeled data for enhancing model generalization. This is particularly crucial in the medical domain where labeled data is scarce. Experiments in the BTCV and MSD datasets demonstrate that LoGoNet surpasses the baselines, achieving superior segmentation accuracy. In the analysis section, we reported numerous experiments. We particularly showed that the combination of LoGoNet with pretraining further enhances accuracy, and the utilization of masked data in pretraining framework significantly boosts the model performance.



\bibliographystyle{ACM-Reference-Format}
\bibliography{references}

\pagebreak

\section*{Appendix}
\label{sec:appendix}

The following section presents a more detailed description of our feature extractor, ULKANet. Then, we provide the details of our experiments, including the configurations of the baseline models, our pre-training algorithm, and our model architecture. We continue with a description of each used dataset, and finally, we conclude the article by reporting an additional qualitative experiment.


\begingroup 
\setlength{\tabcolsep}{4pt} 

\section{Detailed Architecture of ULKANet}
\label{appendix_ulkanet}

\begin{figure}[ht!]
  \centering
  \includegraphics[scale=0.9]{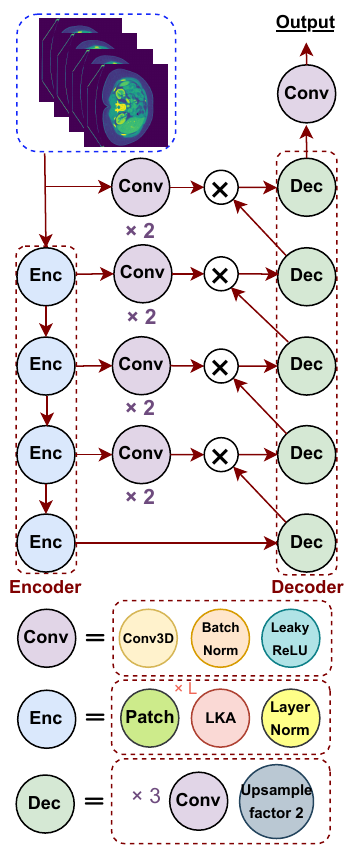}
  \caption{Architecture of our feature extractor (ULKANet). The numbers next to some of the components indicate a sequence of the depicted component with the specified length.}
  \label{fig:ULKA_model}
\end{figure}

Figure \ref{fig:ULKA_model} illustrates our feature extractor. This feature extractor is structured into two main components: an encoder and a decoder. The encoder is comprised of a series of blocks, each consisting of a recurring sequence of three essential elements: a patch embedding component, which you can find the algorithm of this component in the algorithm \ref{alg:patch}, a set of transformer-like modules employing the LKA technique (The number of these modules in the sequence is represented as $L$), and a layer normalization component. The LKA component contains two crucial parts, first attention, which we describe in part \ref{sub-sec:ulkanet}, and the MLP part, which you can find in the algorithm \ref{alg:MLP}; also, the algorithm of LKA part is available in the algorithm \ref{alg:LKA}. This architecture has been meticulously designed to process and extract crucial input data features effectively. The patch embedding operation transforms the input into a feature vector with a dimension of $dim$. Additionally, we incorporate a $Conv$ block, which encompasses three layers: a $Conv3D$ layer, batch normalization, and the $LeakyRelu$ activation function.

Furthermore, the presence of a decoder block denoted as $Dec$ in Figure \ref{fig:ULKA_model} is a crucial element. This block consists of three $Conv$ blocks and an upsampling layer, which upscales the input by a factor of $2$. This comprehensive structure enables our model to efficiently handle the input data and extract meaningful features for further processing.

\begin{algorithm}
\caption{Patch Embedding Pseudo Code}
\label{alg:patch}
\begin{algorithmic}[1]
\Procedure{PatchEmbed3D}{$X$, $dim$, $patchSize$, $inputChannel$, $stride$} 
    \Comment{Input: $X$ is the input tensor, and $dim$ is embed dimension}
    \State $projection \gets Conv3D(inputChannel, dim, kernel=patchSize, stride=stride, padding=patchSize//2)$
    \State $X \gets projection(X)$
    \State $B, C, D, H, W \gets X.Shape$
    \State $X \gets BatchNorm(X)$
    \State $X \gets X.flatten(2).transpose(1, 2)$
    \State \textbf{Return} X, D, H, W
\EndProcedure
\end{algorithmic}
\end{algorithm}

\begin{algorithm}
\caption{Pseudo Code of LKA Block}
\label{alg:LKA}
\begin{algorithmic}[1]
\Procedure{LKA}{$X$, $dim$, $H$, $W$, $mlpRatio$} 
\Comment{Input: $X$ is the input tensor. $dim$, $H$, and $W$ are the dimensions of the input tensor.}
    \State $B, N, C \gets X.shape$
    \State $X \gets X.permute(0, 2, 1).view(B, C, dim, H, W)$
    \State $X \gets BatchNorm(X)$
    \State $attentionValue \gets attentionFunction(X)$ \Comment{The attention function is described before in the part \ref{sub-sec:ulkanet}}
    \State $X \gets X + attentionValue$
    \State $X \gets BatchNorm(X)$
    \State $mlpValue = MLP(X, dim, mlpRatio \times dim)$
    \State $X \gets X + mlpValue$
    \State $X \gets X.view(B, C, N).permute(0, 2, 1)$
    
    \State \textbf{Return} X
\EndProcedure
\end{algorithmic}
\end{algorithm}

\begin{algorithm}
\caption{Pseudo Code of MLP Block}
\label{alg:MLP}
\begin{algorithmic}[1]
\Procedure{MLP}{$X$, $inSize$, $hiddenSize$, $outSize$} 
\Comment{Input: $X$ is the input tensor.}
    \State $fc1 \gets Conv3d(inSize, hiddenSize, kernel=1)$
    \State $X \gets fc1(x)$
    \State $X \gets GELU((X)$
    \State $dwconv3d \gets Conv3d(inSize, inSize, kernel=3)$ 
    \State $X \gets dwconv3d(X)$
    \State $X \gets GELU((X)$
    \State $fc2 =\gets Conv3d(hiddenSize, outSize, kernel=1)$

    \State \textbf{Return} X
\EndProcedure
\end{algorithmic}
\end{algorithm}



\section{Complementary Implementation Details}

\label{app:imp_det}

\begin{table*}[ht]
\begin{tabular}{c|ccc|ccc|ccc|ccc}
\hline
\multirow{2}{*}{\textbf{Layer Number}} & \multicolumn{3}{c|}{\textbf{1}}                                                         & \multicolumn{3}{c|}{\textbf{2}}                                                         & \multicolumn{3}{c|}{\textbf{3}}                                                         & \multicolumn{3}{c}{\textbf{4}}                                                          \\ \cline{2-13} 
                                       & \multicolumn{1}{c|}{\textbf{L}} & \multicolumn{1}{c|}{\textbf{dim}} & \textbf{mlpRatio} & \multicolumn{1}{c|}{\textbf{L}} & \multicolumn{1}{c|}{\textbf{dim}} & \textbf{mlpRatio} & \multicolumn{1}{c|}{\textbf{L}} & \multicolumn{1}{c|}{\textbf{dim}} & \textbf{mlpRatio} & \multicolumn{1}{c|}{\textbf{L}} & \multicolumn{1}{c|}{\textbf{dim}} & \textbf{mlpRatio} \\ \hline
\textbf{Normal}                        & 3                               & 64                                & 8                 & 4                               & 128                               & 8                 & 6                               & 256                               & 4                 & 3                               & 512                               & 4                 \\
\textbf{Large}                         & 3                               & 96                                & 8                 & 3                               & 192                               & 8                 & 24                              & 384                               & 4                 & 3                               & 768                               & 4                 \\ \hline
\end{tabular}

  \caption{The number of LKA modules in each encoder block and mlpRatio for each encoder layer, as well as the embedding dimensions of the Patch Embedding module for the regular and the large variants of our model.}
  \label{tab:version_info}
\end{table*}

\small
\begin{table*}[ht!]
  \centering
    \begin{tabular}{c|ccccccccccccc}
    \hline
          & Spl & RKid & LKid & Gall & Eso & Liv & Sto & Aor & IVC & Veins & Pan & Rad & Lad \\
   \hline
    DiNTS & .937$\pm$.02 & .934$\pm$.00 & .930$\pm$.02 & .788$\pm$.02 & .770$\pm$.00 & .960$\pm$.00 & .774$\pm$.02 & .904$\pm$.02 & .866$\pm$0.23 & .751$\pm$.02 & .813$\pm$.02 & .670$\pm$.02 & .711$\pm$.01 \\
    LoGoNet & .958$\pm$.02 & .949$\pm$.00 & .947$\pm$.01 & .818$\pm$.02 & .786$\pm$.00 & .969$\pm$.01 & .880$\pm$.02 & .912$\pm$.01 & .865$\pm$.01 & .769$\pm$.02 & .821$\pm$.02 & .726$\pm$.01 & .698$\pm$.01 \\
    \hline
    \end{tabular}%
      \caption{Comparison of Performance Metrics (\textit{Mean $\pm$ Standard Deviation}) for Various Methods Across Different Organs}
  \label{tab:yourlabel}%
\end{table*}%

Our model architecture has incorporated four encoder blocks, a feature in both the standard and the larger variants. However, it's important to note that our model is flexible and can seamlessly adapt to the use of varying numbers of encoder layers. The primary distinction between the regular and large models lies in the number of transformer modules within each block and the dimensions of the internal embedding vectors.

To provide a comprehensive understanding, Table \ref{tab:version_info} presents a detailed comparison between our standard model and its larger counterpart. It's noteworthy that, despite any variations, the size of the embedding vectors for each patch module and the mlpRatio remains consistent across all encoder blocks.

This structural consistency ensures that the essential characteristics of the model components are preserved, facilitating ease of integration and adaptability. Whether opting for the standard or larger version, users have the freedom to fine-tune the model's performance by adjusting the number of encoder layers to suit their specific requirements. This flexibility is a key advantage of our model, allowing for versatility in handling diverse applications and tasks.

In the implementation of the local strategy within LoGoNet, a pivotal decision was made to partition each image tensor into $N=8$ segments. While this approach offers advantages in enhancing local processing capabilities, it concurrently introduces a significant surge in the number of trainable parameters. In addressing this challenge, a thoughtful strategy has been employed within the local section of LoGoNet.

Specifically, in the local processing segment of LoGoNet, a judicious selection has been made to utilize only two encoder blocks, in contrast to the four blocks employed in the global section, as previously mentioned. This intentional divergence in the number of encoder blocks between the local and global sections serves to strike a balance between computational complexity and model expressiveness.

By limiting the local section to two encoder blocks, we manage to mitigate the potential escalation in trainable parameters, thereby optimizing the trade-off between computational efficiency and model performance. This strategic choice is rooted in a nuanced understanding of the interplay between local and global processing within the overall architecture of LoGoNet.

In essence, our design rationale carefully tailors the number of encoder blocks in each section to the specific demands of local and global processing, ensuring a harmonious integration that optimally leverages the strengths of both approaches. This meticulous consideration of architectural choices reflects our commitment to achieving a well-balanced and efficient model in LoGoNet.

\section{Complementary Results} \label{sec:com_results}

The information pertaining to pre-training is encapsulated in Table \ref{tab:pre-train-info}. To train the pre-trained model, we leveraged the $AdamW$ optimizer and fine-tuned the process by configuring specific parameters. In particular, we assigned values of $0.1$ and $0.7$ to $\phi_1$ and $\phi_2$ respectively. Additionally, the sequence of distorted images, denoted as $M$, was set to $5$. 

\begin{table}[ht]
\begin{tabular}{c|ccc}
\hline
\textbf{Hyperparameter}  & \textbf{M = 3} & \textbf{M = 5} & \textbf{M = 7} \\ \hline
\textbf{$\phi_1$ = 0.1} & .835  & \textbf{.850}  & {\ul.847}  \\
\textbf{$\phi_1$ = 0.2} & .838  & {\ul.847}  & .840  \\
\textbf{$\phi_1$ = 0.3} & .841  & .843  & .838  \\ \hline
\end{tabular}

  \caption{Hyperparameter tuning for sequenced mask image length (M) and rate of sampled images ($\phi_1$): A detailed exploration of hyperparameter variations to optimize key aspects of our experimental setup. Result is for BTCV dataset and ULKANet model.}
  \label{tab:hyperparameter}
\end{table}

Table \ref{tab:hyperparameter} presents the outcome of selecting hyperparameter values, with results obtained from the BTCV dataset using the ULKANet model. This tabulated information sheds light on the meticulous decision-making process involved in determining specific values for key hyperparameters, providing valuable insights into our experimental configuration.

Our observations reveal that augmenting both the values of $M$ (length of sequenced mask images) and $\phi_1$ (rate of sampled images) results in an increased rate of masked images. However, this heightened rate poses challenges for our model, making it more intricate to exploit dependencies between successive slices for effectively capturing information related to missing voxels. This delicate interplay between hyperparameters emphasizes the necessity of finding an optimal balance to enhance model performance, as an excessive increase in masked images may impede the model's ability to leverage contextual dependencies within the data.

Regarding statistically significant tests, Our model underwent rigorous training, leveraging the BTCV dataset five times and the MSD dataset twice, ensuring robustness and reliability. To comprehensively evaluate our model's performance, we present the average Dice accuracy and standard deviation on the BTCV dataset. Table \ref{tab:yourlabel} shows \textit{Mean $\pm$ Standard Deviation} for our method compared to the best baseline (DiNTS) across different organs. Upon scrutinizing the table, a notable observation emerges; for instance, a one-tail t-test conducted on the 'Gall' class yields a calculated t-value of $t(8) = 2.599$, corresponding to a p-value of $0.016$. Our model demonstrates statistical significance over the best baseline in five classes at an alpha level of $0.05$ and nine classes at an alpha level of $0.10$, elucidating its superior performance across multiple organ segments.

\end{document}